\documentclass{article}

\usepackage[nonatbib, final]{nips_2017}

\usepackage[utf8]{inputenc} 
\usepackage[T1]{fontenc}    
\usepackage{hyperref}       
\usepackage{url}            
\usepackage{booktabs}       
\usepackage{nicefrac}       
\usepackage{microtype}      

\usepackage{textcomp} 
\usepackage{graphicx, caption} 

\usepackage[bitstream-charter]{mathdesign}
\usepackage[T1]{fontenc}

\def\supsub#1#2{\rlap{\textsuperscript{#1}}\sub{#2}}

\def\sub#1{\textsubscript{#1}}

\title{Ablation of a Robot's Brain: \\ Neural Networks Under a Knife}

\author{
  Peter E. Lillian, Richard Meyes, Tobias Meisen \\
  Institute of Information Management in Mechanical Engineering, RWTH Aachen University \\
  Dennewartstr. 27, 52064 Aachen, Germany \\
   \texttt{p@peterlillian.com}, \texttt{\{richard.meyes, tobias.meisen\}@ima-ifu.rwth-aachen.de} \\
}

\begin{document}

\maketitle

\begin{abstract}
It is still not fully understood exactly how neural networks are able to solve the complex tasks that have recently pushed AI research forward. We present a novel method for determining how information is structured inside a neural network. Using \textit{ablation} (a neuroscience technique for cutting away parts of a brain to determine their function), we approach several neural network architectures from a biological perspective. Through an analysis of this method's results, we examine important similarities between biological and artificial neural networks to search for the implicit knowledge locked away in the network's weights. 
\end{abstract}

\section{Introduction}
The 21st century has seen the rapid advancement of machine learning. Novel breakthroughs in deep learning with diverse architectures such as convolutional or recurrent networks have enabled unheard-of results in a wide variety of tasks. \cite{KriSut12Imagenet,DBLP:journals/corr/Graham14a,DBLP:journals/corr/DonahueHGRVSD14,DBLP:journals/corr/ZhaoMGL15} More specifically, work in Reinforcement Learning (RL) has finally created agents that are able to learn and interact with their environment to accomplish a goal---without needing an explicit hand-crafted feature representation. \cite{mnih2013playing}

But, even with advancements in the understanding of neural networks, the field at large has yet to arrive at a "comprehensive understanding" of how these fascinating models are organized---how they actually operate. \cite{Shwartz-ZivT17} This is the cause of the so-called \lq{}black box\rq{} problem that has plagued researchers since the discovery of backpropagation, preventing the interpretation of results. The acquiescence in this phrase discourages study into the organizational structure and is not conducive to furthering our understanding of these models and improving their usefulness. \cite{DBLP:journals/corr/AlainB16}

In the traditional computer, there are dedicated components for the storage of knowledge, memory of instructions, and processing of the two. Traditional programs can take advantage of these components, and this makes it easy to understand what an algorithm does and how it works. Information here is stored \textit{explicitly}: we can find exactly where the bits exist in physical space. In a neural network though, all of these functions (memory, et cetera) are learned, saved, and utilized in its weights. Information here is stored \textit{implicitly}: we are not certain how it is distributed across neurons or where it exists, and we cannot access it directly. How then, can we fully understand the emergent organization of neural networks?

We can take a look at another \lq{}black box\rq{} medium for inspiration: the human brain. \cite{becker11neuroblackbox} Though biological neural networks are not fully understood, two key characteristics are well known and may shed some light on neural networks.

First, the brain is incredibly robust and resilient to damage. Destroy many neurons or even large areas and in many cases it will more or less continue to function. \cite{cicchetti2006multiple} In fact, it is actually difficult to damage the brain in a way that makes it impossible to perform most tasks. The most prominent medical example of this is the case of Phineas Gage, a railroad foreman involved in an accident where a thick metal rod 3.2cm in diameter passed completely through his left frontal lobe. \cite{Damasio1102} After several years of recovery, he was able to live a relatively normal life as a stagecoach driver. Human evolution in uncertain and violent environments has necessitated this kind of robustness. 

Second, the brain tends not only to create an internal mapping for external inputs, like it does in sensory and motor cortices, but also to specialize each area to accomplish a different task. \cite{grodd2001sensorimotor} Essentially, the brain's representation of knowledge (or at least the way this is stored) physically reflects external input---this idea is called functional specialization or localization of function. In the motor and sensory cortices, each physical part of the body is mapped to a different physical part of the brain, creating a kind of internal body \lq{}topographic map\rq{}. \cite{grodd2001sensorimotor} This is the case for the brain's sensory inputs as well as its motor outputs and is an important feature of biological networks.

Neuroscientists have researched these characteristics in many ways, but one effective method is an \textit{ablation study}. Ablation is the selective removal or destruction of tissue for the purpose of understanding the function of the tissue in question---starting with early brain experiments, it has been used in research for over two hundred years. \cite{carlson2009psychology} More recently, ablation studies have been used to accurately pinpoint which neural substrates are involved in specific types of memory. \cite{MURRAY199613} 

These studies have been applied to neural networks, but in a different way than in neuroscience. Current ablation studies in the literature focus on tweaking the layers and changing the network's structure or its implementation. \cite{DBLP:journals/corr/BrockLRW16a} This, however, more closely resembles a parameter search than a biological ablation. Recently, ablation has also been used to analyze network-wide characteristics and make inferences about network properties. \cite{DBLP:journals/corr/abs-1803-06959} When it has been used to understand a network's learned information, it has been applied to the input image itself rather than the network in an attempt to determine what is recognized about the image. \cite{IshaqDeepFish}

In our previous work, we examined the wire-loop problem: the task of robotically guiding a metal loop down a curving wire without one physically contacting the other. \cite{MEYES2017} We improved the results of this study in another recent paper (allowing the model to generalize much better and deal with on-the-fly changes in wire shape). \cite{MEYES2018} We accomplished this by incorporating an actor-critic architecture trained with Deep Deterministic Policy Gradient (DDPG) based upon Lillicrap et al. (2015). \cite{DBLP:journals/corr/LillicrapHPHETS15} A persistent problem we encountered, though, was the model's tendency to learn inaccurate, highly variable motion. On a non-curving path, the model rarely learns to move straight forward. This is true even though a reward of -1 is given for each motion without progress.

With all this in mind, our goal for this study was to search for parallels between the emergent organization of biological and artificial neural networks. By approaching from a conceptual perspective, we show that these very different networks do share important characteristics and that further research into their similarities is needed.

\section{Background and Methods}
The main idea underpinning this study is that by performing ablation on neural networks we will discover a similar type of robustness to damage and localization of function that is present in biological networks. While there are of course massive low-level differences between these two types of networks, the correspondence between their larger learned structures is unknown. Consider the mammalian brain: one could imagine splitting the motor cortex into many small groups of neurons and iteratively removing/replacing each one (as in an ablation study). By testing the output of the network (in the biological case, behavior is the output) after each ablation and looking for differences, we could determine the function of each ablated group.

\begin{figure}[!phtb]
  \centering
  \includegraphics[width=\textwidth{}]{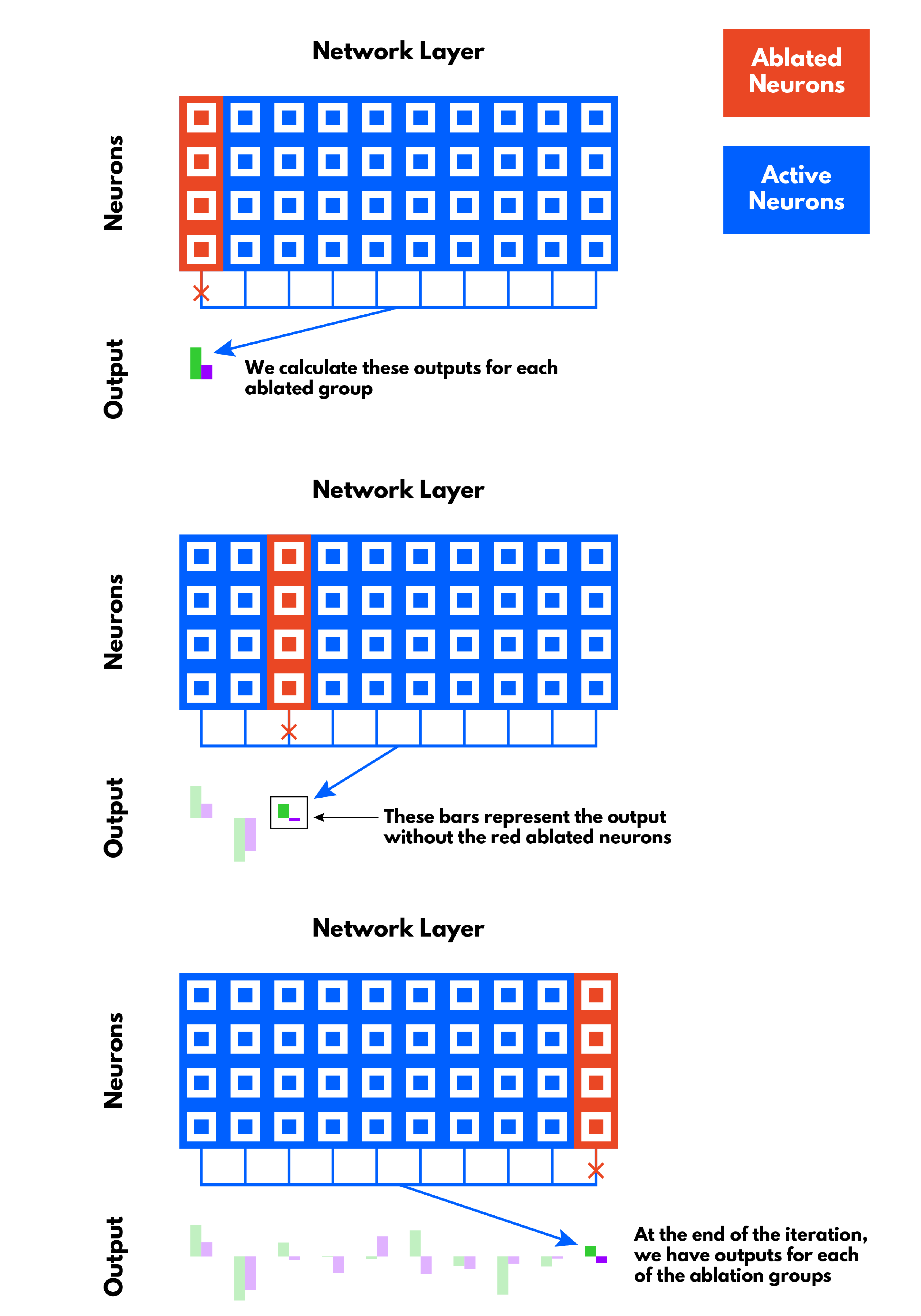}
  \caption{As each ablation group is removed, the output without that group is saved. After ablating each group, we have a list of outputs showing how the network changes when its parts are removed. Only one group is ablated at a time.}
  \label{fig:ablation_groups}
\end{figure}

Neural networks may not seem to conform to the biological expectation that related functions are positioned closely together, and this may be true for networks trained on arbitrarily sorted features. Convolutions performed on images, however, necessarily preserve spatial information, which suggests that neighboring neurons will represent similar information.



Reinforcement Learning is a subfield of machine learning focused on creating an agent that can learn strategies to solve tasks by interacting with its environment in discrete time. After observing a state \textit{s\textsubscript{t}}, the model predicts an action \textit{a\textsubscript{t}} and receives a reward \textit{r\textsubscript{t}}. The model's goal is to find a policy \textit{$\pi$} that maximizes future reward, and it does this through an random exploration process. The actor network in our model, once trained, predicts the action that gives the greatest future reward.

Our actor network, modified slightly from our recent paper, down-samples the input using three consecutive convolutional layers of sizes 5x5, 5x5, and 3x3 with 30, 15, and 10 filters respectively (each activated by ReLU and followed by a 2x2 max-pooling layer). Following these layers are a fully-connected network, consisting of layers of output size 400 and 200, followed by three layers of output size 1 (also activated by ReLU). All these layers utilize a dropout rate of 0.2 during the training phase. Together, the three outputs predict each component \textit{a\supsub{i}{t}} of the action \textit{a\textsubscript{t}}, made up of a \lq{}forward\rq{} longitudinal motion, \lq{}left\rq{} lateral motion, and \lq{}left rotational motion.\rq{} In the later two actions, left is motion in the positive direction and right is in the negative direction. \cite{MEYES2018} These components form an action space in ([0, 1], [-1, 1], [-1, 1]) respectively, multiplied by scale factors 3cm, 3cm, 90\textdegree{}.


Before starting the study, we compute the baseline output for each network and each image using a forward pass for later comparison with the ablated networks. To actually perform the ablation study, we focused on the second-to-last layer of each network. We divided its neurons into 10 groups. 
Then we iterate through each group, setting all weights and biases associated with those neurons to zero and running a forward pass to calculate the new outputs. Now we can compare the baseline outputs to the outputs with each ablated group removed. (See Figure ~\ref{fig:ablation_groups} for a visual explanation.) We trained the model from scratch in 5 trials. We define a trial as training of the model from scratch until it is able to successfully navigate a wire.

Splitting the network into groups in this way might be arbitrary, but by comparing the outputs of each ablated group and checking which ones change, we can identify the average function of the neurons in each group. This allows us to find regions of the networks across multiple trials that perform similar functions.

For our actor network, we compared different trials to determine if groups of neurons with similar functions arose when the network was trained on completely different wire paths. For our purpose, we trained our RL model from scratch in 5 of these trials. We compared these trials by treating the outputs calculated on each image (the actions that the network predicts) as a set of features to be analyzed. We will call this the \textit{expanded action space}. A given network generates 2 useful actions for each image: \lq{}left lateral\rq{} and \lq{}left rotational\rq{} because the forward longitudinal action is highly constant no matter the input. With 8 images per category and 3 categories, this expanded space contains a total of 48 features per ablation network. Because the layer of our network undergoing ablation was subdivided into 10 neuron groups, each trial gives 10 data points of 48 features each. These 48 features, compared to the baseline features, represent how removing this group of neurons changes the network---and it can reveal what knowledge was stored there. For example, if a given region of a network was involved in turning the robot left, then the output features of the images of left-turning wires should change upon ablation of that region. From this point, we can normalize the data from each trial and use K-means clustering to group these ablated networks across different trials. Points nearby in this expanded action space represent ablations that created similar changes in output, so the clusters group similar parts of the network across different trials. Two different trials may not create the same internal structure, but through this method we can determine ablation group analogues.

\section{Results}

\begin{figure}[!htbp]
  \centering
  \includegraphics[width=\textwidth{}]{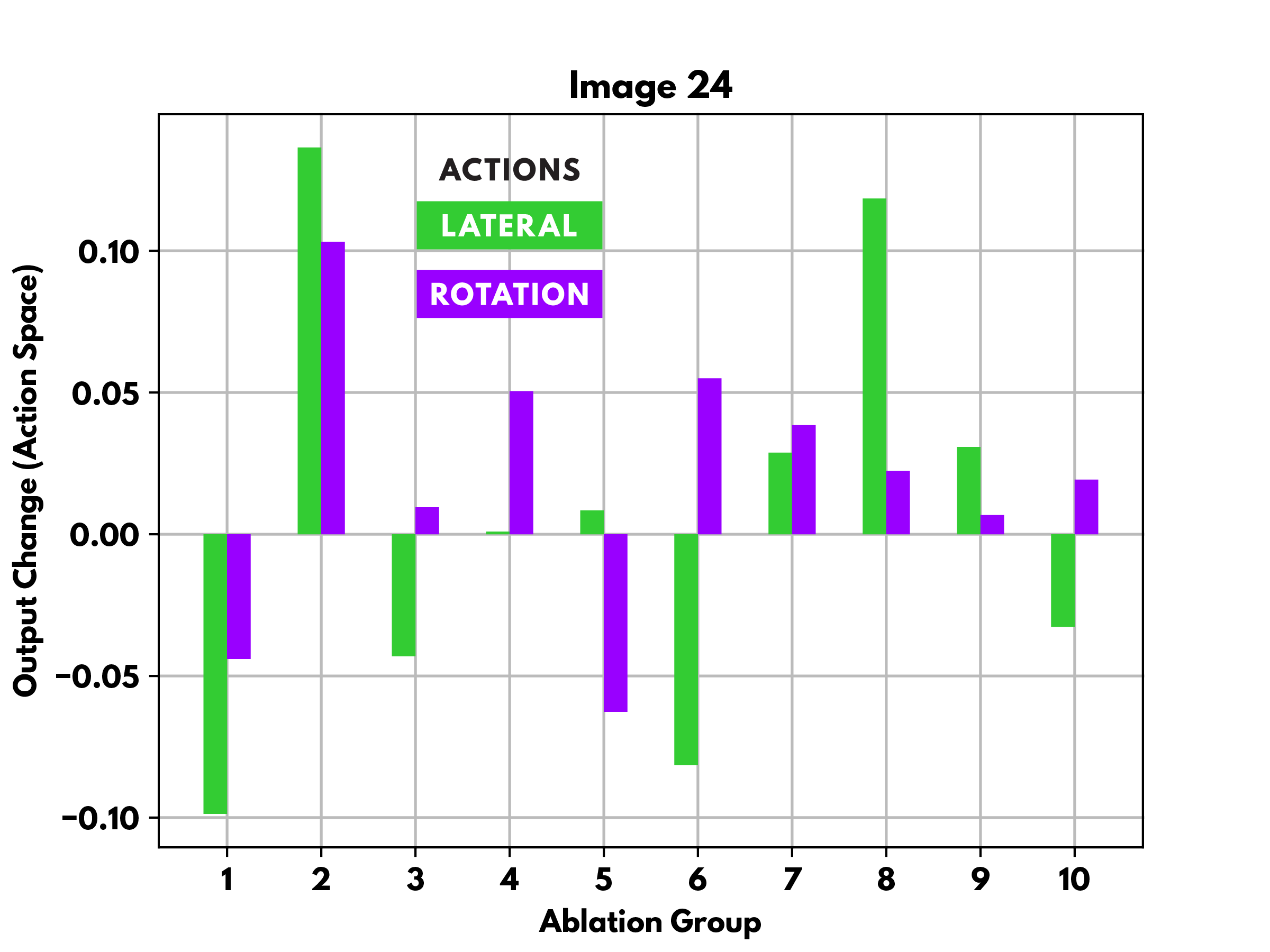}
  \caption{A typical network's (using our model) ablation results (how its output changed) for an image---our method matches each ablation group with its counterparts in the other trials. This data makes up part of the expanded action space. Additionally, we have omitted the longitudinal action due to its highly constant value.}
  \label{fig:ablation_results_24}
\end{figure}

In each trial, for each ablated group, we found that network output was shifted in one direction regardless of image, and only a few ablation groups actually changed the action by a large margin. Because each trial is randomly initialized, their final structures differed greatly. Using silhouette scoring, the clustering algorithm consistently found that the data is best fit by 6 clusters: we can extrapolate from this that the neurons serve 6 loosely-defined roles. In order to discover these roles, we must examine the nodes in each cluster.



\begin{figure}[!htbp]
  \centering
  \includegraphics[width=0.6\textwidth{}]{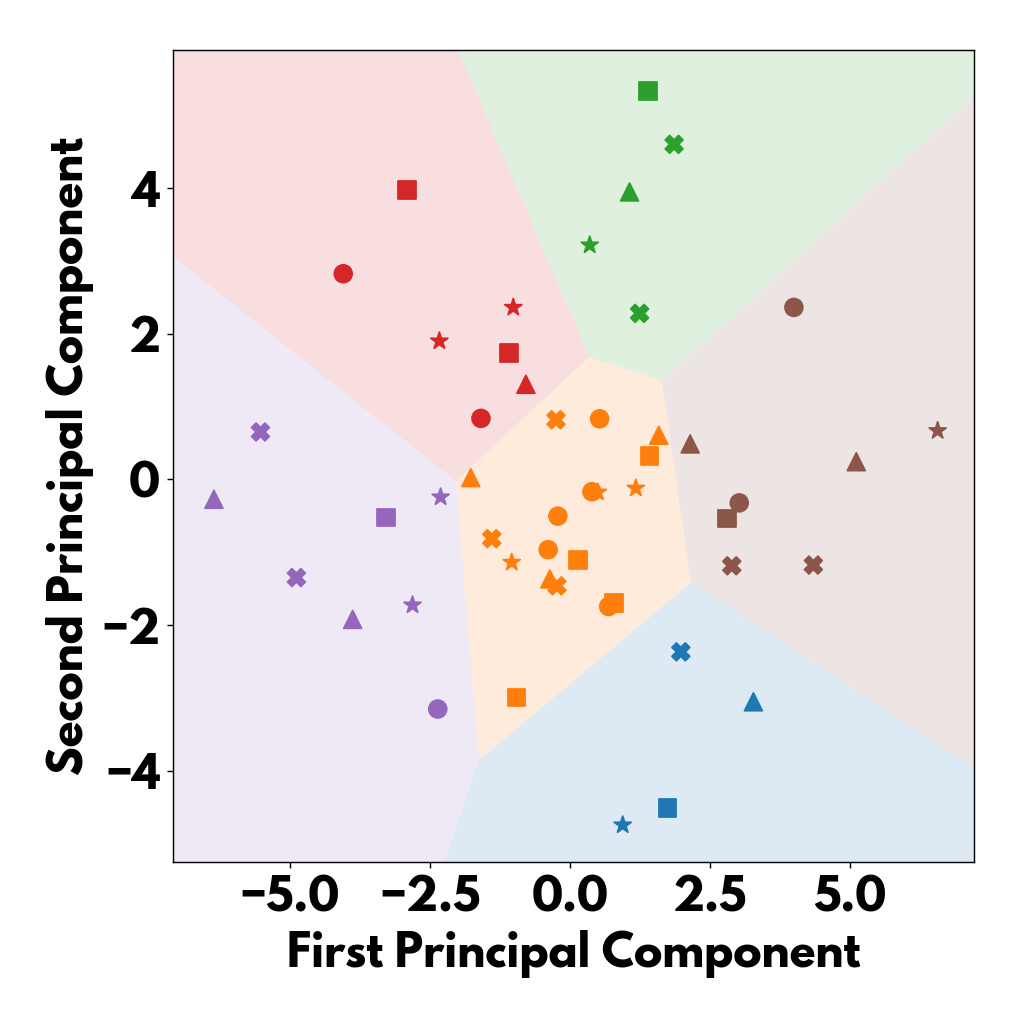}
  \caption{The expanded action space of our model reduced to two dimensions by Principal Component Analysis (PCA). Colors represent the clusters found by K-means and each shape represents a different trial. (The data was clustered before PCA was applied.) K-means Voronoi boundaries are also shown.}
  \label{fig:pca_action_space}
\end{figure}

\begin{figure}[!htbp]
  \centering
  \includegraphics[width=0.6\textwidth{}]{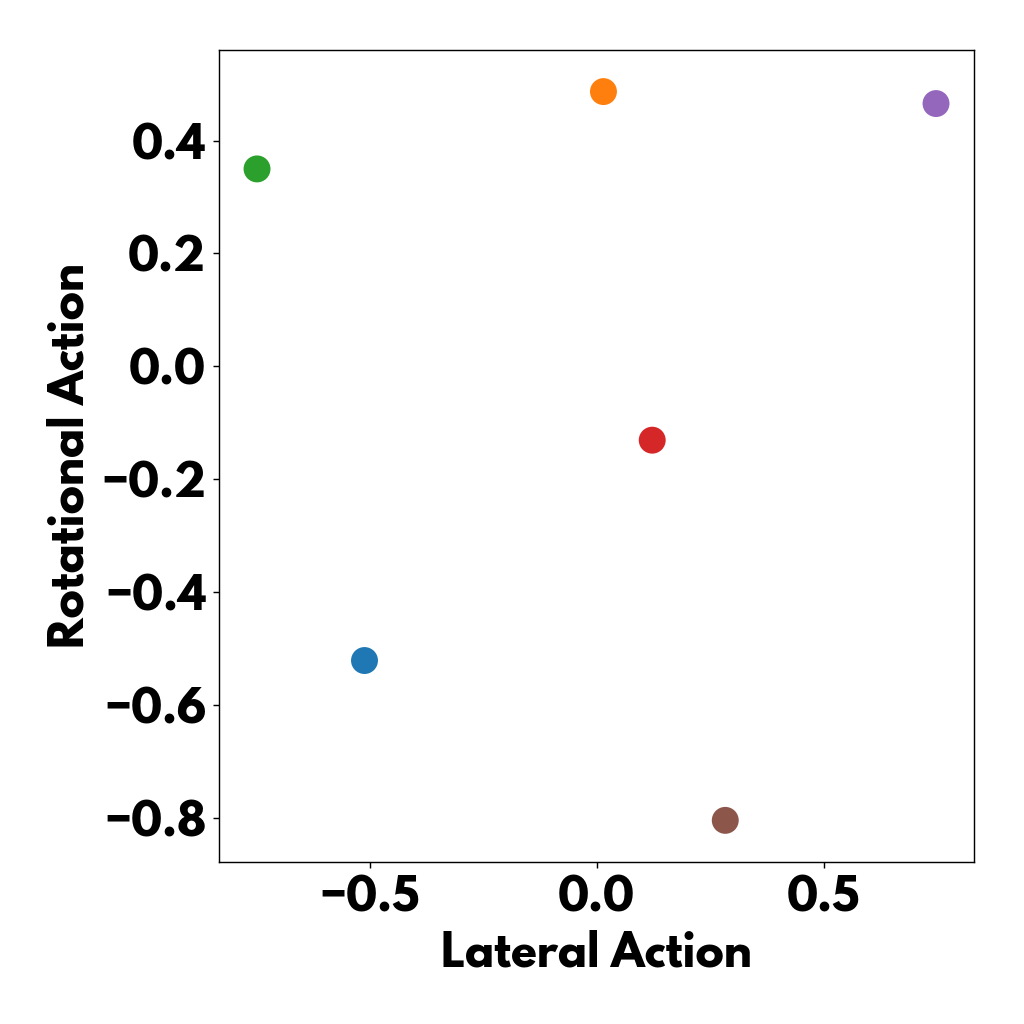}
  \caption{For each cluster in Figure ~\ref{fig:pca_action_space}, the average change in action upon that cluster's ablation is shown.}
  \label{fig:avg_clusters}
\end{figure}

For both types of models, Principal Component Analysis (PCA) was used for visualization of these high-dimensional expanded action spaces. Figure ~\ref{fig:pca_action_space} shows the first and second Principal Components for the expanded action space created by ablations of our actor model. We clustered using K-means, iteratively calculating the average silhouette score for a given number of clusters \textit{n}. For out model, this number \textit{n} was varied from 2 to 12 and it was found that \textit{n}=6 clusters consistently gave the best silhouette score of 0.309. While all the rest of the scores were closely grouped within 1.5 IQR of the median, \textit{n}=6 was a far outlier: 4.9 IQR above this median. For this reason \textit{n} was set to 6 for the final clustering, as this result implies that there are 6 types of ablation groups. Once the results were clustered, we found that the trials' ablation groups were well distributed across the expanded action space. All clusters contained one ablation group from 4 out of the 5 trials, and half contained them from all 5. This shows that each time the model is trained from scratch in a new trial, there are regions that perform the same functions as analogous regions in other trials.

Upon further investigation into these 6 clusters, it becomes apparent that each 'type' of ablation group has a different average function. (We use the word average here because our method looks at groups of spatially adjacent neurons, which may not all have the same function.) For example, when ablation groups of the blue and green types from Figure ~\ref{fig:pca_action_space} are removed, the network's output shifts strongly in the right lateral direction. A similar case holds for the other groups of neurons; we observe that each type affects the output in one consistent way, whereas ablation of that group affects the output in the opposite way. In the same way as the human brain, this network exhibits clear localization of function. Conversely, we also see the networks' resistance to damage here. The groups work together, so the ablation of a group involved in moving right will only be knocking out one pathway. We can see from, for example, Figure ~\ref{fig:ablation_results_24} that while each ablation group has a specific role, the removal of one group never completely destroys the ability of the network to complete a given action.


We understand that our results are merely the first step in a thorough investigation of the connections between biological and artificial neural networks. 
One particular issue with our study is the granularity of the ablations: further studies must look at these similarities at different scales.
Additionally, future investigations into these ideas should use larger datasets. Ours was limited by our use of reinforcement learning 
and reliance on a model trained robotically in realtime.

\section{Summary and Discussion}
In this paper, we explored critical links between biological and artificial neural networks: both in how they arrange themselves and store information. By ablating neural networks trained to navigate a wire-loop and analyzing how the output of each network changed, we were able to examine these links and demonstrate their strength. 

We plan to further investigate these links to make more progress toward better theory. This may even inspire more efficient neural network implementations that exploit these insights.

Clearly, at the low level, both types of networks are fundamentally different. This makes it all the more interesting that these networks resemble one another in the ways we have shown. By no means is this paper trying to be a perfect analysis of these resemblances: further study, including a more granular analysis will be necessary to more fully understand what is going on. We believe that this is necessary due to the field's engineering approach to problem solving. While this approach has created much in terms of useful methods, ultimately the field is still lacking a low-level conceptual understanding of artificial neural networks. The creation of this kind of framework would enable a move beyond educated trial and error.

\small
\bibliographystyle{ieeetr}
\bibliography{references}

\end{document}